\pdfoutput=1
\documentclass[conference]{IEEEtran}
\IEEEoverridecommandlockouts
\usepackage{cite}
\usepackage{amsmath,amssymb,amsfonts}
\usepackage{graphicx}
\usepackage{textcomp}
\usepackage{xcolor}

\usepackage{algorithm}
\usepackage{algorithmicx}
\usepackage{algpseudocode}
\usepackage{booktabs}
\usepackage{multirow}
\usepackage{gensymb}

\def\BibTeX{{\rm B\kern-.05em{\sc i\kern-.025em b}\kern-.08em
    T\kern-.1667em\lower.7ex\hbox{E}\kern-.125emX}}
\begin{document}

\title{Learning Multi-Modal Brain Tumor Segmentation from Privileged Semi-Paired MRI Images with Curriculum Disentanglement Learning 
\thanks{$ ^{*} $ corresponding author}
}

\author{\IEEEauthorblockN{Zecheng Liu}
\IEEEauthorblockA{\textit{School of Computer Science} \\
\textit{and Engineering} \\
\textit{South China University of Technology}\\
Guangzhou, China \\
scutzcliu@gmail.com}
\and
\IEEEauthorblockN{Jia Wei$ ^{*} $}
\IEEEauthorblockA{\textit{School of Computer Science} \\
\textit{and Engineering} \\
\textit{South China University of Technology}\\
Guangzhou, China \\
csjwei@scut.edu.cn}
\and
\IEEEauthorblockN{Rui Li}
\IEEEauthorblockA{\textit{Golisano College of Computing} \\
\textit{and Information Sciences} \\
\textit{Rochester Institute of Technology}\\
Rochester, NY, USA \\
rxlics@rit.edu}
}

\maketitle

\begin{abstract}
Due to the difficulties of obtaining multimodal paired images in clinical practice, recent studies propose to train brain tumor segmentation models with unpaired images and capture complementary information through modality translation. However, these models cannot fully exploit the complementary information from different modalities. In this work, we thus present a novel two-step (intra-modality and inter-modality) curriculum disentanglement learning framework to effectively utilize privileged semi-paired images, i.e. limited paired images that are only available in training, for brain tumor segmentation. Specifically, in the first step, we propose to conduct reconstruction and segmentation with augmented intra-modality style-consistent images. In the second step, the model jointly performs reconstruction, unsupervised/supervised translation, and segmentation for both unpaired and paired inter-modality images. A content consistency loss and a supervised translation loss are proposed to leverage complementary information from different modalities in this step. Through these two steps, our method effectively extracts modality-specific style codes describing the attenuation of tissue features and image contrast, and modality-invariant content codes containing anatomical and functional information from the input images. Experiments on three brain tumor segmentation tasks show that our model outperforms competing segmentation models based on unpaired images.
\end{abstract}
 

\section{Introduction}
\label{sec:introduction}

Automatic and accurate segmentation of brain tumors is essential to diagnosis and treatment planning. It requires precisely detecting both the location and extent of a tumor~\cite{gms:13,rnmb:13,wczc:14,xl:07}. However, the uncertainty of tumorous shape, size and location poses a unique challenge, especially in infiltrative tumors such as gliomas~\cite{mjb:14,ppas:16,wk:08}. A common solution is to integrate information acquired from multimodal paired MRI, since different pulse sequences (modalities) of MRI provide complementary information of brain tumors from multi-perspectives~\cite{glhg:19,yyfyjcyz:20,zhzhhy:21}. 

\begin{figure}[t]
	\centering
	\includegraphics[width=0.9\columnwidth]{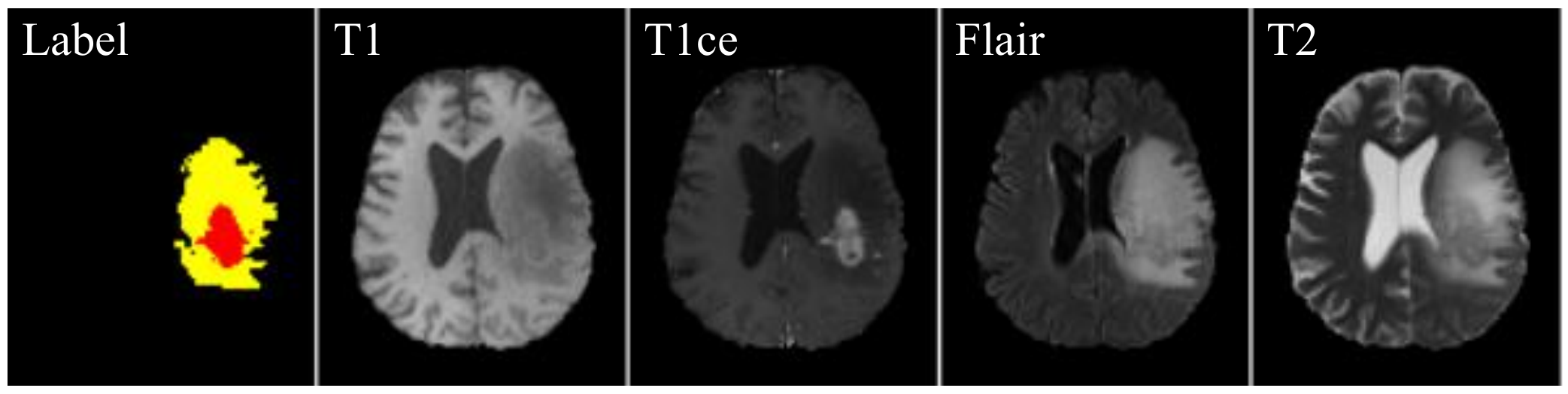}
	\caption{Different brain tumor information of the same subject can be detected from different sequences of brain MRI. In the Label image, yellow area indicates a whole peritumoral edema, and red area indicates a tumor core.}
	\label{fig1}
\end{figure}

As shown in Fig.~\ref{fig1}, T1ce (contrast enhanced T1-weighted) highlights tumors without peritumoral, but the image contrast of the whole peritumoral edema is enhanced in T2 (native T2-weighted) and Flair (T2 Fluid Attenuated Inversion Recovery)~\cite{zrc:19}. Although these methods show promising performances, they require paired data in both training and test, as illustrated in Fig.~\ref{fig21}(a). This hinders their applicability in clinical practice, when only unpaired or missing modality images are available.

To alleviate the above problem, a multimodal unpaired learning method was proposed for medical image segmentation~\cite{vprlarrg:18,ywwmt:19}, as shown in Fig.~\ref{fig21}(b). For instance, Yuan {\em et al}.~\cite{ywwmt:19} proposed a two-stream translation and segmentation unified attentional generative adversarial network. The model is trained with unpaired data and performs predictions for unpaired images by capturing and calibrating complementary information from translation to improve segmentation. However, without the supervision of paired images, the quality of image translation tends to be poor, especially for brain tumor areas. This is because the method cannot effectively exploit the complementary information from different modalities such as varying shapes of brain tumors.

\begin{figure*}[t]
	\centering
	\includegraphics[width=0.85\textwidth]{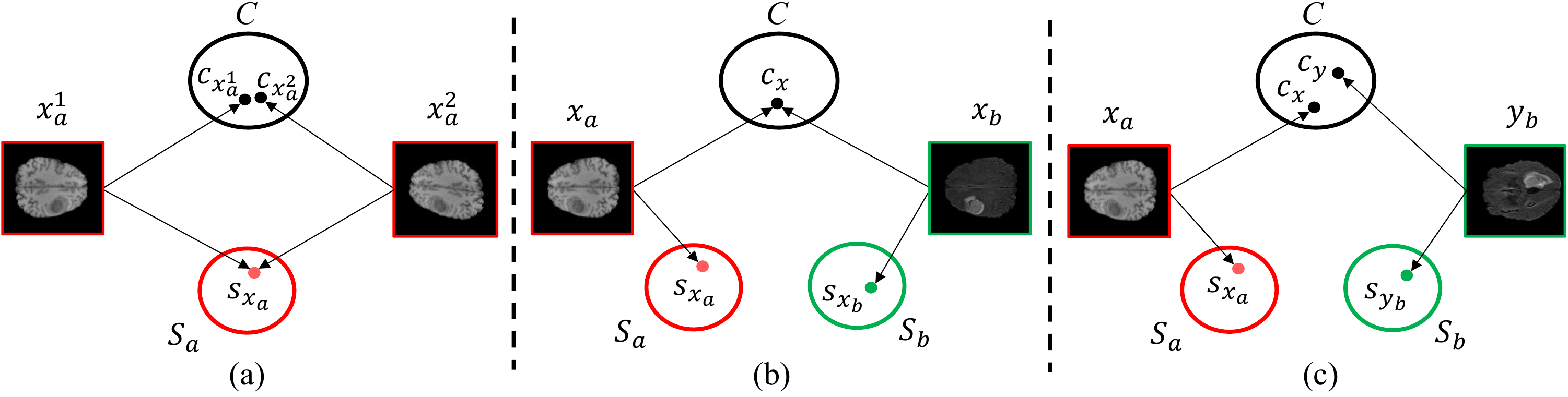} 
	\caption{Different feature disentanglement learning schemes. (a) Style-consistent learning scheme. Style-consistent images obtained by horizontal flip ($ x^{1}_{a} $) and elastic deformation ($ x^{2}_{a} $) are given in this example. (b) Paired inter-modality learning scheme. (c) Unpaired inter-modality learning scheme.}
	\label{fig22}
\end{figure*}

\begin{figure}[t]
	\centering
	\includegraphics[width=0.9\columnwidth]{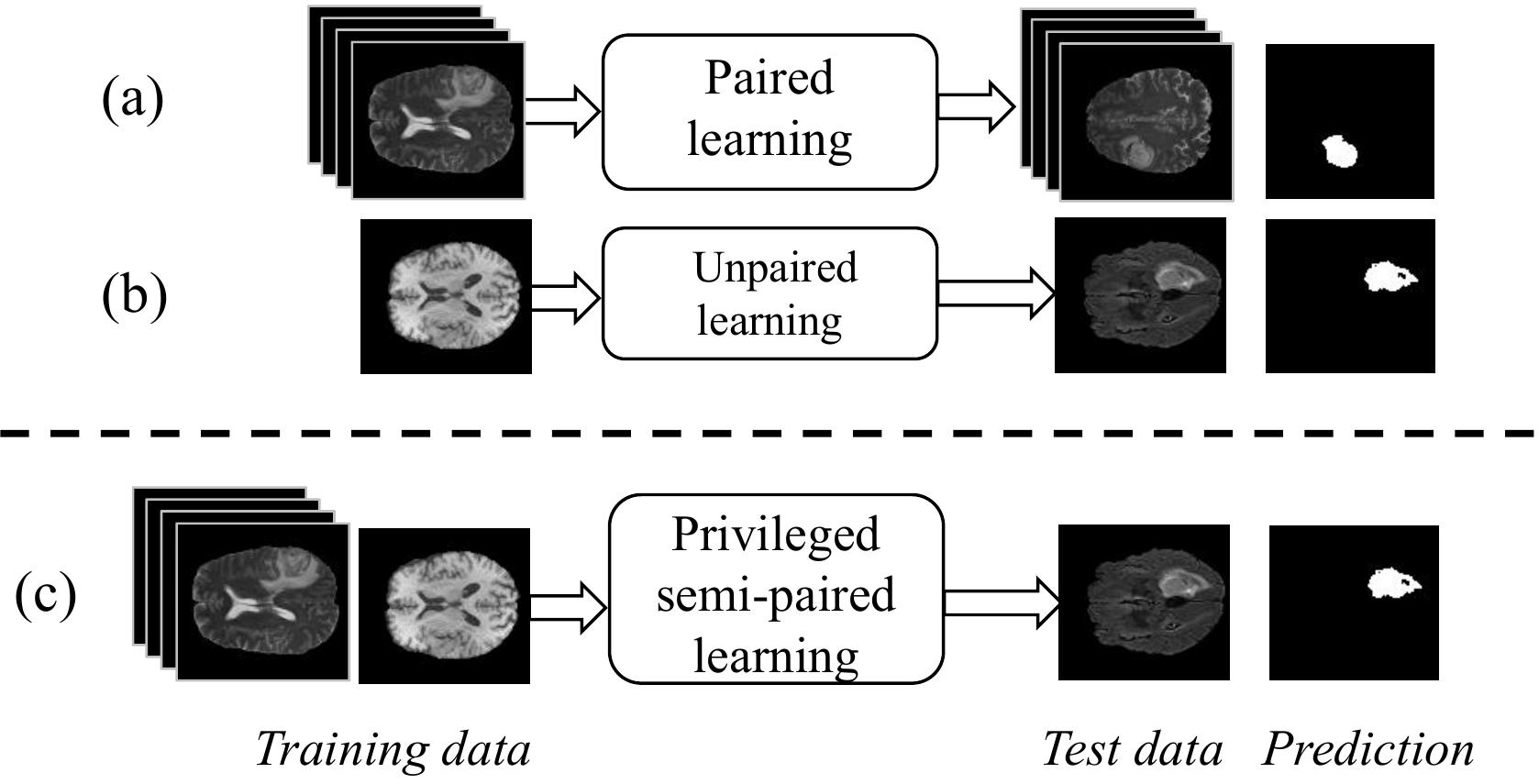}
	\caption{Illustration of (a) Paired learning, (b) Unpaired learning, and (c) Privileged semi-paired learning frameworks.}
	\label{fig21}
\end{figure}

In this work, we propose a novel two-step curriculum disentanglement learning method to leverage privileged semi-paired images for brain tumor segmentation, as shown in Fig.~\ref{fig21}(c), when limited paired images are only available in training. Compared to previous feature disentanglement learning models (DRIT~\cite{lthsy:18}, MUNIT~\cite{hlbk:18}), the proposed model focuses on effective separation of style and content. As shown in Fig.~\ref{fig22}(c), the previous models only use unpaired inter-modality learning scheme for training. Unpaired images $ x_{a} $ (from subject $ x $ of modality $ a $) and $ y_{b} $ (from subject $ y $ of modality $ b $) are mapped into the same content space but different style spaces. However, there are no specific constraints for these spaces, so the disentanglement mapping tends to incur large variations, which results in the problem of ambiguous separation between style and content. Ouyang et al. \cite{oap:21} investigate the problem. They require that the content representations from the same patient with different modalities should be as similar as possible. However, under the constraint, the style and other task-unrelated components (e.g., noise and artifacts) tend to corrupt the content representations, which fails to reduce the ambiguity of the content and style. On the contrary, we contend that different modalities of a given patient essentially reflect the inherent anatomy of the patient, which is consistent even though its appearance may be diverse across different modalities. Therefore, to solve this problem, we propose a curriculum disentanglement learning strategy with two steps:

(1) In the first step, as shown in Fig.~\ref{fig22}(a), we generate two style-consistent images $ x^{1}_{a} $ and $ x^{2}_{a} $ from the same original image $ x_{a} $ with different image processing methods (such as horizontal flip and elastic deformation). We then define a style consistency loss to map the images to the same point $ s_{x_{a}} $ in the style space $ S_{a} $. 

(2) The second step is consist of reconstruction, unsupervised/supervised translation, and segmentation based on unpaired and paired inter-modality learning schemes. The unpaired inter-modality learning scheme, as shown in Fig.~\ref{fig22}(c), maps unpaired images $ x_{a} $ and $ y_{b} $ obtained from different subjects to different points in the content space $ C $. The paired inter-modality learning scheme, as shown in Fig.~\ref{fig22}(b), maps paired images $ x_{a} $ and $ x_{b} $ obtained from the same subject $ x $ to the same point $ c_{x} $ in the content space $ C $ with a content consistency loss. 

Through the two steps, our proposed method can separate modality-specific style codes and modality-invariant content codes from the input images. In particular, the modality-specific style codes describe attenuation of tissue features and image contrast, and the modality-invariant content codes contain consistent inherent anatomical and functional information. The effective disentanglement of the two codes is critical for brain tumor segmentation.

The contributions of this paper are as follows:
\begin{itemize}
	\item We propose a multi-task disentanglement learning framework which leverages privileged semi-paired images for brain tumor segmentation. A content consistency loss and a supervised translation loss are proposed to integrate complementary information.
	\item We propose a two-step (intra-modality and inter-modality) curriculum disentanglement learning strategy to train our model, so that we can effectively separate the style and content of input images.
	\item We provide qualitative and quantitative performance evaluations on brain tumor segmentation tasks with BraTS2020~\cite{brats:20} and BraTS2018~\cite{brats:18}. The results show the superiority of our method over the state-of-art unpaired medical image segmentation methods.
	\item We further demonstrate the superior performance of our model on a modality translation task, which suggests that the segmentation can be improved by fully exploiting complementary information.
\end{itemize}

\begin{figure*}[t]
	\centering
	\includegraphics[width=0.8\textwidth]{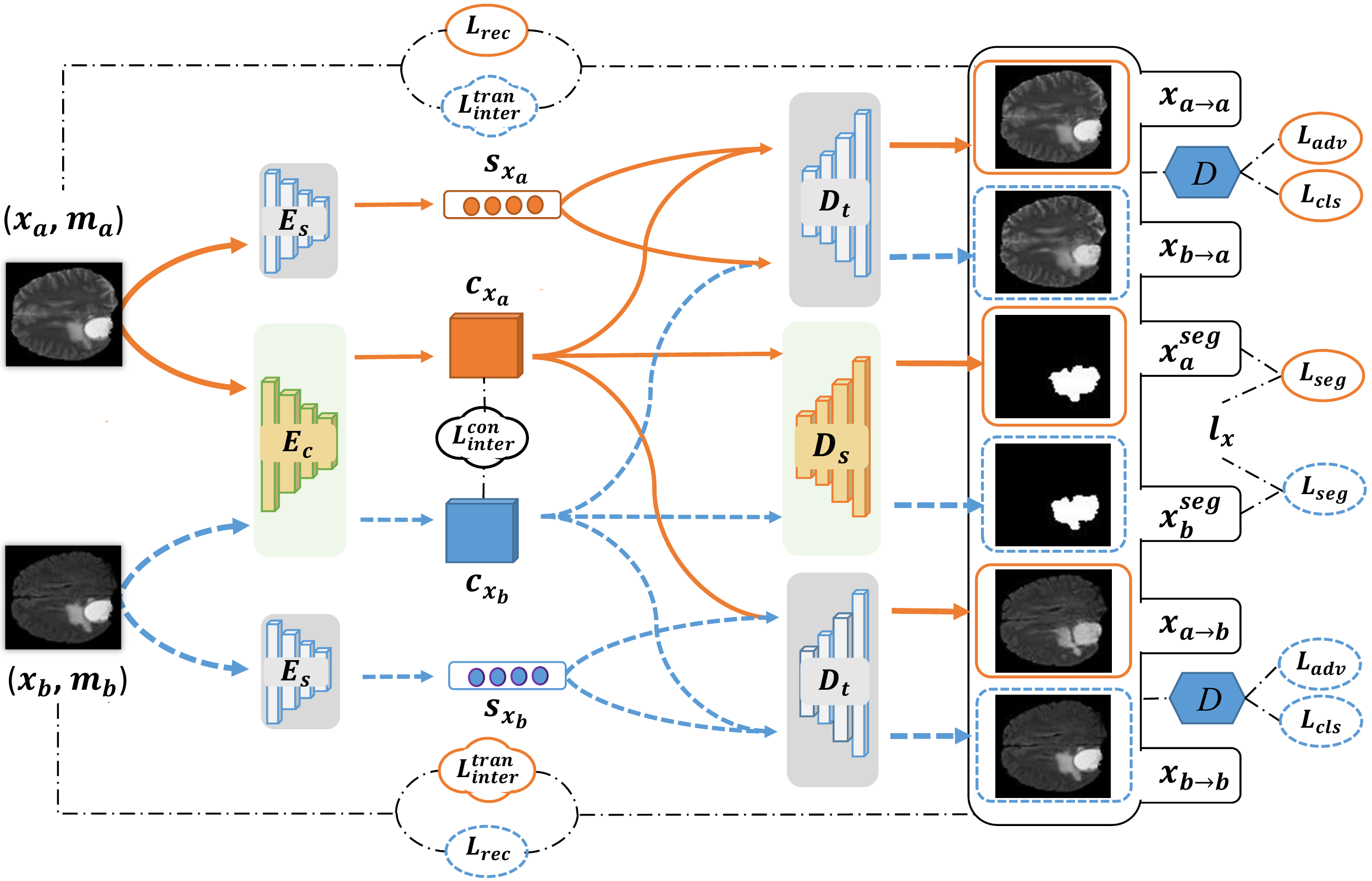} 
	\caption{Illustration of the proposed framework. Paired images for the inter-modality learning scheme are depicted in this example. All the networks are unified, including the content and style encoders ($ E_{c},E_{s} $), the translation and segmentation decoders ($ D_{t},D_{s} $), and the discriminator $ D $. The data stream of images $ x_{a} $ and $ x_{b} $ are drawn as solid orange arrows and dotted blue arrows. Losses are computed by the corresponding generated images and segmentation map (orange solid box for $ x_{a} $, blue dotted box for $ x_{b} $). Note that image modality translation loss $ L^{tran}_{inter} $ and content consistency loss $ L^{con}_{inter} $ are only applied to paired images in the inter-modality learning scheme.}
	\label{fig2}
\end{figure*}

\section{Related work}
The methods for multimodal brain tumor segmentation can be broadly separated into two categories: segmentation through paired learning and segmentation through unpaired learning.
\subsection{Segmentation through paired learning}
Multiple imaging modalities have been widely used in medical image segmentation due to its ability to provide complementary information to reduce information uncertainty. During the past few years, most researches focused on the multi-modal fusion strategies, such as input-level fusion and layer-level fusion. These methods either concatenate multi-modality images as multi-channel inputs~\cite{zdlwt:18} or fuse the features from different network trained by different modality~\cite{dgylda:18}. The improvement in the accuracy of brain tumor segmentation relies on the exploitation of complementary information. However, these methods rely on paired data in both training and test, and it hinders their applicability in clinical practice, where only unpaired or missing modality images are available.

Recently, to mitigate performance degradation when inferencing, medical image segmentation with missing modality has been extensively studied. The most popular approach is to fuse the available modalities in a latent space to learn a shared feature representation for segmentation. A variable number of input modalities are mapped to a unified representation by computing the first and second moments~\cite{hgcb:16}, mean function~\cite{kjj:19}, or fusion block~\cite{cdjcqh:19,zcvr:21}. Moreover, Shen {\em et al}.~\cite{szwxpthsmcwx:21} utilize synthesizd images as multi-channel inputs to obtain shared representation for segmentation with a multimodal image completion and segmentation disentanglement network called ReMIC. Furthermore, Chen {\em et al}.~\cite{cdj:21} proposed a privileged knowledge learning framework with the ``Teacher Student" architecture. Privileged information is transferred from a multimodal teacher network to a unimodal student network for unpair images. However, the method also requires a large amount of paired images for multimodal teacher network training. Our method, instead, utilizes privileged semi-paired images, where only limited paired images are available for training.

\subsection{Segmentation through unpaired learning}
For medical image segmentation, in order to utilize all available data for training even when the images are unpaired, an X-shaped multiple encoder-decoder networks was proposed~\cite{vprlarrg:18}. The model extracts modality-independent features to improve segmentation accuracy by sharing the last layers of the encoders. Information from one modality is captured in the shared network to improve the performance of segmentation task on another modality. Furthermore, Yuan {\em et al}.~\cite{ywwmt:19} proposed a two-stream translation and segmentation network called UAGAN. The network captures inferred complementary information from modality translation task to improve segmentation performance. The above methods do not require any paired images, and utilize easily accessible unpaired images for training, instead. However, these methods cannot integrate complementary information without paired images. On the contrary, our method can effectively leverage complenentary information of limited number of paired images by encoding them into a modality-invariant content space through content consistency constraint and supervised translation for brain tumor segmentation.

\section{Methodology}
\subsection{Proposed model}
As shown in Fig.~\ref{fig2}, we use paired images as an example to illustrate our framework. Given images $ x_{a} $, $ x_{b} $ from the same subject and different modalities. We adopt one-hot vectors to represent their modality label and expands them to the same image size, denoted as $ m_{a} $ and $ m_{b} $. Given the depth-wise concatenation $ (x_{a}, m_{a}) $ and $ (x_{b}, m_{b}) $, our goal is to trian a single generator $ G $ that can simultaneously accomplish the following tasks: (1) Reconstructing the input images $ x_{a} $ and $ x_{b} $ as $ x_{a \to a} $ and $ x_{b \to b} $ respectively. (2) Translating $ x_{a} $ of modality $ m_{a} $ to the corresponding output image $ x_{a \to b} $ of modality $ m_{b} $, and $ x_{b} $ of modality $ m_{b} $ to $ x_{b \to a} $ of modality $ m_{a} $. (3) Generating brain tumor segmentation masks $ x^{seg}_{a} $ and $ x^{seg}_{b} $ of the input images $ x_{a} $ and $ x_{b} $, respectively. We denote it as $ G((x_{a}, m_{a}),(x_{b}, m_{b})){\to}(x_{a \to a},x_{a \to b},x^{seg}_{a},x_{b \to b}, x_{b \to a},$ $ x^{seg}_{b}) $. The architecture of our model is composed of two modules described below.

We design the generator $ G $ with four shared networks ($ E_{s} $,$ E_{c} $,$ D_{s} $,$ D_{t} $) based on feature disentanglement. Given the input concatenation $(x_{a}, m_{a})$, the network $ E_{s} $ generates its style code $ s_{x_{a}} $ which is a vector with dimension $ n_{s} $, and the network $ E_{c} $ generates its content code $ c_{x_{a}} $ which is a feature map, denoted as $ E_{s}((x_{a}, m_{a})){\to}s_{x_{a}} $ and $ E_{c}((x_{a}, m_{a})){\to}c_{x_{a}} $. Similarly, $ s_{x_{b}} $ and $ c_{x_{b}} $ are also obtained from the input $ (x_{b}, m_{b}) $, denoted as $ E_{s}((x_{b}, m_{b})){\to}s_{x_{b}} $ and $ E_{c}((x_{b}, m_{b})){\to}c_{x_{b}} $. Then, we perform image reconstruction, translation, and segmentation base on these disentangled representations. For image reconstruction, given the content code and the style code obtained from the same input image, the decoder $ D_{t} $ generates the corresponding reconstruction image, denoted as $ D_{t}(c_{x_{a}}, s_{x_{a}}){\to}x_{a \to a} $ and $ D_{t}(c_{x_{b}}, s_{x_{b}}){\to}x_{b \to b} $. For image translation, given the content code and the style code obtained from different input images, the decoder $ D_{t} $ translates the source image of one modality (corresponding to the content code) to the target image of the other modality (corresponding to the style code). We denoted it as  $ D_{t}(c_{x_{a}}, s_{x_{b}}){\to}x_{a \to b} $ and $ D_{t}(c_{x_{b}}, s_{x_{a}}){\to}x_{b \to a} $. For image segmentation, given the content code, the decoder $ D_{s} $ generates a binary mask to identify and highlight the tumor area of the corresponding input image, denoted as $ D_{s}(c_{x_{a}}){\to}x^{seg}_{a} $ and $ D_{s}(c_{x_{b}}){\to}x^{seg}_{b} $.

The probability distributions produced by the discriminator $ D $ distinguish whether the generated images from $ G $ are real or fake, and determine which modality they are from.

\subsection{Constructing the objective function}
Our loss function consists of three parts: (1) common losses for both curriculum disentanglement learning steps; (2) losses for intra-modality disentanglement step; (3) losses for inter-modality disentanglement step. For simplicity, we only describe the losses for image $ x_{a} $, since the loss function for $ x_{b} $ is the same. Alg.~\ref{alg:algorithm} summarizes the overall procedure of the curriculum disentanglement learning.

\begin{algorithm}[tp]
	\caption{The curriculum disentanglement learning}
	\label{alg:algorithm}
	
	\textbf{\textit{Training input}}: intra-modality augmented style-consistent images ($ x^{1}_{a} $, $ x^{2}_{a} $)$ _{1} $, $\ldots$, ($ x^{1}_{a} $, $ x^{2}_{a} $)$ _{i} $, paired inter-modality images ($ x_{a} $, $ x_{b} $)$ _{1} $, $\ldots$, ($ x_{a} $, $ x_{b} $)$ _{j} $, and unpaired inter-modality images ($ x_{a} $, $ y_{b} $)$ _{1} $, $\ldots$, ($ x_{a} $, $ y_{b} $)$ _{k} $ \\
	\textbf{\textit{Training output}}: generator $ G $
	
	\begin{algorithmic}[1] 
		\While{not converged}  \textcolor{blue}{// In the first step}
		\State \textcolor{blue}{// Style-consistent pattern}
		\State Updata $ G $ and $ D $ using Eq.~\eqref{loss:G_intra} and Eq.~\eqref{loss:D}
		\EndWhile
		\While{not converged} \textcolor{blue}{// In the second step}
		\State \textcolor{blue}{// Paired inter-modality pattern}
		\State Updata $ G $ and $ D $ using Eq.~\eqref{loss:G_inter} and Eq.~\eqref{loss:D}
		\State \textcolor{blue}{// Unpaired inter-modality pattern} 
		\State Updata $ G $ and $ D $ using Eq.~\eqref{loss:G} and Eq.~\eqref{loss:D}
		\EndWhile		
	\end{algorithmic}
	
	\textbf{\textit{Test input}}: unpaired images $ x_{1} $, $\ldots$, $ x_{n} $ \\
	\textbf{\textit{Test output}}: segmentation results  $ x^{seg}_{1} $, $\ldots$, $ x^{seg}_{n} $
	\begin{algorithmic}[1] 
		\State Calculate $ {\forall}r $, $ x^{seg}_{r} \leftarrow G((x_{r},m_{r})) $ \\
		\textcolor{blue}{// $ m_{r} $ is the modality label of $ x_{r} $}
	\end{algorithmic}
	
\end{algorithm}

\subsubsection{Common losses} Losses for both curriculum disentanglement learning steps include an adversarial loss, a modality classification loss, a reconstruction loss, and a segmentation loss.

Adversarial loss: To minimize the difference between the distributions of generated images and real images, we define the adversarial loss as:
\begin{equation}
	\begin{aligned}
		\begin{split}
			L_{adv}=&\mathbb{E}_{x_{a}}[\log D_{src}(x_{a})]+\frac{1}{2}\mathbb{E}_{x_{a \to a }}[\log(1-D_{src}(x_{a \to a }))] \\
			&+\frac{1}{2}\mathbb{E}_{x_{a \to b}}[\log(1-D_{src}(x_{a \to b}))]	
			\label{loss:adv}
		\end{split}
	\end{aligned}
\end{equation}
$ D_{src} $ denotes probability distributions, given by discriminator $ D $, of real or fake images~\cite{cckhkc:18}. The discriminator $ D $ maximizes this objective to distinguish between real and fake images, while the generator $ G $ tries to generate more realistic images to fool the discriminator.

Modality classification loss: To allocate the generated image to correct modality, the modality classification loss is imposed to $ G $ and $ D $. It contains two terms: modality classification loss of real images which is used to optimize $ D $, denoted as $ L^{r}_{cls} $, and the loss of fake images used to optimize $ G $, denoted as $ L^{f}_{cls} $.
\begin{equation}
	L^{r}_{cls}=\mathbb{E}_{x_{a}}[-\log{D_{cls}(m_{a}|x_{a})}]
	\label{loss:dcls}
\end{equation}
\begin{equation}
	\begin{split}
		L^{f}_{cls}=\frac{1}{2}\mathbb{E}_{x_{a \to a}}[-\log(D_{cls}(m_{a}|x_{a \to a}))] \\
		+\frac{1}{2}\mathbb{E}_{x_{a \to b}}[-\log(D_{cls}(m_{b}|x_{a \to b}))]
	\end{split}
	\label{loss:gcls}
\end{equation}
Here, $ D_{cls} $ represent the probability distributions over modality
labels and input images~\cite{cckhkc:18}.

Reconstruction loss: To prevent the omission of detailed information, we employ the reconstruction loss to constrain the recovered images: 
\begin{equation}
	L_{rec}=\mathbb{E}_{x_{a}, x_{a \to a}}[\|x_{a \to a}-x_{a}\|_{1}]
\end{equation}

Segmentation loss: The segmentation loss is a dice loss:
\begin{equation}
	L_{seg}=-\frac{2\sum^{N}_{i=1}l^{i}_{x}x^{i}_{seg}}{\sum^{N}_{i=1}(l^{i}_{x}l^{i}_{x}+x^{i}_{seg}x^{i}_{seg})+\epsilon }
\end{equation}
Here, $ l^{i}_{x} $, $ x^{i}_{seg} $ denote ground truth and prediction of voxel $ i $, respectively. The $\epsilon=1e^{-7}$ is a constant for numerical stability.

Objective functions: By combining the above losses together, our common objective functions are as follows:
\begin{equation}
	L_{D}=-L_{adv}+L^{r}_{cls}
	\label{loss:D}
\end{equation}
\begin{equation}
	L_{G}=L_{adv}+L^{f}_{cls}+\lambda_{rec}L_{rec}+\lambda_{seg}L_{seg}
	\label{loss:G}
\end{equation}
where $ \lambda_{rec} $ and $ \lambda_{seg} $ are hyperparameters to control the relative importance of reconstruction loss and segmentation loss. In addition, we use L2 norm regularization to constrain the style codes to encourage a smooth space and minimise the encoded information~\cite{lm:15}.

\subsubsection{Curriculum losses for the first step}
In the first intra-modality disentanglement step, we train the model with style-consistent learning scheme shown in Fig.~\ref{fig22}(a). The augmented images are created by the following six image processing methods: (1) horizontal flip, (2) vertical flip, (3) rotate random angle (0\degree$\sim$360\degree), (4) zoom in to random ratios (0.8$\sim$1.2), (5) elastic deformation~\cite{rfb:15}, and (6) shift a random distance (0px$\sim$20px) in all directions. For each original image $ x_{a} $ in training data (both paired and unpaired image), we randomly use two methods to obtain two style-consistent images denoted as $ x^{1}_{a} $ and $ x^{2}_{a} $. Note that original image can belong to any modality. Let $ a $ indexes a modality, and $ s_{x^{1}_{a}} $ and $ s_{x^{2}_{a}} $ denote the style codes of $ x^{1}_{a} $ and $ x^{2}_{a} $, respectively. We define a style consistency loss to constrain $ s_{x^{1}_{a}} $ and $ s_{x^{2}_{a}} $ to be similar:

\begin{equation}
	L^{sty}_{intra}=\mathbb{E}_{(s_{x^{1}_{a}},s_{x^{2}_{a}})}[\|s_{x^{1}_{a}}-s_{x^{2}_{a}}\|_{1}]
\end{equation}
In the intra-modality step, the objective function to optimize $ D $ is as in Eq.~\eqref{loss:D}, while the objective function to optimize $ G $ is defined as:
\begin{equation}
	L_{G_{intra}}=L_{G}+\lambda_{sty}L^{sty}_{intra}
	\label{loss:G_intra}
\end{equation}
Here, $ L_{G} $ is as in Eq.~\eqref{loss:G} and the $ \lambda_{sty} $ is the hyperparameter to control the contribution of $ L^{sty}_{intra} $.

\subsubsection{Curriculum losses for the second step}
In the second inter-modality disentanglement step, the training data include both paired and unpaired images from different modalities. 

The objective function for $ D $ in Eq.~\eqref{loss:D} and the objective function for $ G $ in Eq.~\eqref{loss:G} generate different content codes and different style codes for unpaired image ($ x_{a} $, $ y_{b} $).

For paired images ($ x_{a} $, $ x_{b} $), they have the same content codes and different style codes, so we construct a content consistency loss to constrain their content codes:
\begin{equation}
	L^{con}_{inter}=\mathbb{E}_{(c_{x_{a}},c_{x_{b}})}[\|c_{x_{a}}-c_{x_{b}}\|_{1}]
	\label{loss:cc_inter}
\end{equation}
In addition, the image $ x_{a \to b} $ generated from the translation task $ D_{t}(c_{x_{a}}, s_{x_{b}}) $ is expected to be consistent with the image $ x_{b \to b} $ generated from the reconstruction task $ D_{t}(c_{x_{b}}, s_{x_{b}} $), since that $ x_{a} $ and $ x_{b} $ are paired. Meanwhile, $ x_{b \to b} $ is the reconstruction image of $ x_{b} $, $ x_{a \to b} $ is expected to be consistent with $ x_{b} $. Thus, we introduce a translation loss to further constrain $ c_{x_{a}} $ and $ c_{x_{b}} $ as:
\begin{equation}
	L^{tran}_{inter}=\mathbb{E}_{x_{a \to b}, x_{b}}[\|x_{a \to b}-x_{b}\|_{1}]
	\label{loss:t_inter}
\end{equation}
Therefore, the objective function to optimize $ D $ is the same as in Eq.~\eqref{loss:D}, while the objective function to optimize $ G $ is defined as:
\begin{equation}
	L_{G_{inter}}=L_{G}+\lambda_{con}L^{con}_{inter}+\lambda_{tran}L^{tran}_{inter}
	\label{loss:G_inter}
\end{equation}
where, $ L_{G} $ is as in Eq.~\eqref{loss:G} and the $ \lambda_{con} $ and $ \lambda_{tran} $ are hyperparameters to control the contributions of $ L^{con}_{inter} $ and $ L^{tran}_{inter} $.

\section{Experiments and results}
In this section, we first introduce the experimental settings, including datasets, baseline methods, evaluaion metrics, and implementation details. Then, we present and discuss quantitative and qualitative results of our method, including brain tumor segmentation, image translation, ablation study, influence of paired subjects, and disentanglement evaluation.

\begin{table*}[t]
	\centering
	\renewcommand{\arraystretch}{1.2}
	\caption{Performance evaluation for the segmentation task of WT, TC and ET on BraTS2020. A better method has higher Dice (Best highlighted in bold).}
	\setlength{\tabcolsep}{4mm}{
		\begin{tabular}{c|l|cccc|c}
			\toprule[2pt]
			\multicolumn{2}{c|}{Metric} &
			\multicolumn{5}{c}{Dice(\%)$ \uparrow $}\\
			\hline
			\multicolumn{2}{c|}{Modality} &
			T1ce & T1 & T2 & Flair & Aver \\
			\hline
			\multirow{4}{*}{WT} &nnU-Net~\cite{ijfvm:21} 
			& 78.36$\pm$2.23&	74.52$\pm$0.25&	84.18$\pm$1.39&	88.22$\pm$0.29&	81.31$\pm$0.21 \\

			&UAGAN~\cite{ywwmt:19} 
			& 75.56$\pm$1.36&	75.05$\pm$2.40&	82.62$\pm$0.60&	84.53$\pm$0.23&	79.44$\pm$0.06	 \\
			
			&ReMIC~\cite{szwxpthsmcwx:21}
			& 72.18$\pm$1.05&	74.63$\pm$0.18&	76.96$\pm$0.04&	75.37$\pm$3.59&	74.78$\pm$0.67 \\
			
			&Ours
			& \textbf{79.58$\pm$1.13}&	\textbf{77.80$\pm$2.16}&	\textbf{85.66$\pm$0.35}&	\textbf{88.58$\pm$0.08}&	\textbf{82.91$\pm$0.36} \\
			
			\hline
			
			\multirow{4}{*}{TC} &nnU-Net~\cite{ijfvm:21} 
			& 84.78$\pm$1.88&	53.35$\pm$1.33&	66.59$\pm$2.70&	66.63$\pm$0.97&	67.84$\pm$0.78 \\
			
			&UAGAN~\cite{ywwmt:19} 
			& 80.76$\pm$0.65&	58.53$\pm$0.67&	67.99$\pm$0.65&	70.13$\pm$0.90&	69.35$\pm$0.71 \\
			
			&ReMIC~\cite{szwxpthsmcwx:21}
			& 80.68$\pm$3.17&	53.86$\pm$4.72&	62.19$\pm$3.16&	55.14$\pm$0.86&	62.96$\pm$2.98 \\
			
			&Ours
			& \textbf{85.37$\pm$2.05}&	\textbf{62.18$\pm$0.71}&	\textbf{71.68$\pm$1.44}&	\textbf{71.27$\pm$2.80}&	\textbf{72.62$\pm$0.37} \\
			
			\hline
			
			\multirow{4}{*}{ET} &nnU-Net~\cite{ijfvm:21} 
			& 82.09$\pm$1.07&	28.17$\pm$0.30&	45.08$\pm$2.17&	40.03$\pm$5.30&	48.84$\pm$2.06 \\
			
			&UAGAN~\cite{ywwmt:19} 
			& 75.30$\pm$3.01&	33.64$\pm$2.74&	47.19$\pm$3.56&	46.24$\pm$5.53&	50.64$\pm$3.63 \\
			
			&ReMIC~\cite{szwxpthsmcwx:21}
			& 74.98$\pm$0.33&	32.16$\pm$2.26&	39.22$\pm$3.44&	34.68$\pm$0.16&	45.29$\pm$1.34 \\
			
			&Ours
			& \textbf{82.33$\pm$1.04}&	\textbf{37.47$\pm$0.80}&	\textbf{51.61$\pm$3.41}&	\textbf{47.78$\pm$6.36}&	\textbf{54.80$\pm$2.50} \\
			
			\bottomrule[2pt]
			
	\end{tabular} }
	\label{table1}
\end{table*}

\subsection{Experimental settings}
\subsubsection{Datasets} 
To validate the proposed model, we conduct experiments on BraTS2020 \cite{brats:20} and BraTS2018~\cite{brats:18} that consists of 369 and 285 subjects, respectively. Each subject consists of one segmentation mask and four modality scans: T1, T1ce, T2, Flair. There are three mutually inclusive tumor regions:
\textbf{ET} (the enhancing tumor), \textbf{TC} (the tumor core), and \textbf{WT} (the whole tumor). In BraTS2020, we utilize 240 subjects as semi-paired training data, 60 subjects as unpaired validation data, and 69 subjects as unpaired test data. For semi-paired training data, we use 40 of 240 subjects as paired data, while the rest as unpaired data. In BraTS2018, we utilize 180 subjects as semi-paired training data, 50 subjects as unpaired validation data and 55 subjects as unpaired test data. For semi-paired training data, we use 32 of 180 subjects as paired data, while the rest as unpaired data. The subjects are evenly divided between four modalities. For all the images, we resize them to 128$ \times $128 uniformly.

\subsubsection{Baseline methods} 
Segmentation results are evaluated by comparing with the following methods: (1) nnU-Net~\cite{ijfvm:21}, which achieves the best performance in the BraTS2020 competition. Since only limited paired images are available in the training data and all the test data are unpaired, we implement it as unpaired learning (Fig.~\ref{fig21}(b)). The model is trained and tested on four mixed modalities where images are unpaired. (2) UAGAN~\cite{ywwmt:19}, a recently proposed unpaired brain tumor segmentation model, is a two-stream trainslation and segmentation network. Inferred complementary information are captured in the modality translation task to improve segmentation. Since the model is unpaired, we take semi-paired training data as unpaired data for training. (3) ReMIC~\cite{szwxpthsmcwx:21}, a recently proposed image completion and segmentation model for random missing modalities, first achieve image completion, and then concatenate the synthesized modalities as multi-channel inputs to obtain shared representation for segmentation. 

\subsubsection{Evaluation metrics}
We evaluate segmentation performance with dice score (Dice). We compute the metric on every modality, and report average values. In the translation tasks, we use structural similarity (SSIM) as an evaluation metric.

\subsubsection{Implementation details}
The content encoder $ E_{c} $ and the segmentation decoder $ D_{s} $ in the segmentation generator is similar to the U-Net~\cite{rfb:15}. The style encoder $ E_{s} $ and image generation decoder $ D_{t} $ are adapted from~\cite{lthsy:18}. In our experiments, we set $ \lambda_{rec}=50 $ , $ \lambda_{tran}=100 $, $ \lambda_{con}=10 $, $ \lambda_{sty}=10 $, and $ \lambda_{seg}=100 $. The batch size and training epoch are 8 and 50 respectively. We train the model with 20 epochs in the first step and 30 epochs in the second step, which leads to convergence in practice. The dimensionality of the style code is set to $ n_{s}=8 $. All models are optimized with Adam~\cite{kb:14}, and the initial learning rates are $ 1e^{-4} $, $ \beta_{1}=0.9 $, and $ \beta_{2}=0.999 $. The learning rate is fixed in the first 40 epochs, and then linearly declines to $ 1e^{-6} $. All images are normalized to $ [-1,1] $ prior to the training and testing. Our implementation is on an NVIDIA RTX 3090 (24G) with PyTorch 1.8.1.

\begin{table}[t]
	\centering
	\small
	\renewcommand{\arraystretch}{1.2}
	\caption{Performance evaluation of WT segmentation on BraTS2018.}
	\begin{tabular}{l|c|c|c|c|c}
		\hline
		\multicolumn{1}{c|}{Metric} &
		\multicolumn{5}{c}{Dice(\%)$ \uparrow $} \\
		\hline
		\multicolumn{1}{c|}{Modality} &
		T1Gd & T1 & T2 & FLAIR & Aver \\
		\hline
		nnU-Net \cite{ijfvm:21} 
		& 84.10&	76.12&	\textbf{85.91}&	86.16&	83.18\\
		UAGAN \cite{ywwmt:19} 
		& 78.14&	75.61&	80.86&	80.89&	78.95\\
		ReMIC \cite{szwxpthsmcwx:21}
		& 77.97&	72.69&	76.75&	72.23&	74.89\\
		Ours
		& \textbf{86.56}&	\textbf{82.65}&	84.93&	\textbf{87.86}&	\textbf{85.51}\\ 
		\hline
		
	\end{tabular}
	\label{table18}
\end{table}

\begin{table}[t]
	\centering	
	\footnotesize
	\renewcommand{\arraystretch}{1.2}
	\caption{Quantitative evaluations on translated images.}
	\begin{tabular}{l|c|c|c|c|c}
		\hline
		\multicolumn{1}{c|}{Metric} & \multicolumn{5}{c}{SSIM$ \uparrow $}  \\
		\hline
		\multicolumn{1}{c|}{Modality} & T1Gd & T1 & T2 & FLAIR & Aver \\
		\hline
		UAGAN \cite{ywwmt:19}&
		0.5153&	0.4193&	0.3850&	0.4328&	0.4382 \\		
		ReMIC \cite{szwxpthsmcwx:21}&
		0.7205&	\textbf{0.7748}&	0.7579&	0.7061&	0.7398 \\	
		Ours &
		\textbf{0.7741}&	0.7701&	\textbf{0.7905}&	\textbf{0.7671}&	\textbf{0.7754} \\ 
		\hline	
	\end{tabular}
	\label{table4}
\end{table}

\begin{table}[t]
	\centering
	\renewcommand{\arraystretch}{1.2}
	\caption{Performance evaluation of the WT segmentation for ablation study on components. w/o means without.}
	\setlength{\tabcolsep}{2mm}{
		\begin{tabular}{l|cccc|c}
			\hline
			\multicolumn{1}{c|}{Metric} &
			\multicolumn{5}{c}{Dice(\%)$ \uparrow $} \\
			\hline
			\multicolumn{1}{c|}{Modality} &
			T1ce & T1 & T2 & Flair & Aver \\
			\hline
			Ours w/o Cc, T, Cd, F 
			& 76.09&	72.53&	80.04&	84.78&	78.36 \\
			Ours w/o Cc, T, Cd 
			& 75.78&	74.35&	84.04&	87.32&	80.38 \\
			Ours w/o Cc, T
			& 77.88&	76.68&	85.06&	87.97&	81.90 \\
			Ours w/o Cd
			& 77.25&	74.69&	84.46&	87.29&	80.93 \\
			Ours w/o T 
			& 78.50&	77.66&	85.16&	87.95&	82.32 \\
			Ours w/o Cc 
			& 78.76&	77.81&	85.21&	88.15&	82.47 \\
			Ours end-to-end
			& 78.27&	78.44&	84.62&	87.85&	82.30 \\
			Ours 
			& 79.58&	77.80&	85.66&	88.58&	82.91 \\
			\hline
			
	\end{tabular} }
	\label{tableablation}
	
\end{table}

\subsection{Results and analyses}
\subsubsection{Brain tumor segmentation}
We first evaluate the brain tumor segmentation performance of our model on BraTS2020 segmentation tasks (WT, TC and ET). Quantitative results are shown in Table~\ref{table1}. Our model achieves the best overall performance and outperforms the others in all cases. Compared with the best performer of the state-of-the-art methods, our method improves the average dice score from 81.31\% to 82.91\%, 69.35\% to 72.62\% and 50.64\% to 54.80\% on WT, TC and ET segmentation tasks, respectively. In addition, we further evaluate our model on BraTS2018 WT segmentation task. The dice score results are shown in Table~\ref{table18}, our method achieves superior performance in most cases, and improves the average dice score from 83.18\% to 85.51\% compared with nnU-Net. The segmentation results indicate that our method can effectively exploit complementary information by leveraging privileged semi-paired images through the curriculum disentanglement learning model.

\begin{figure*}[t]
	\centering
	\includegraphics[width=0.8\textwidth]{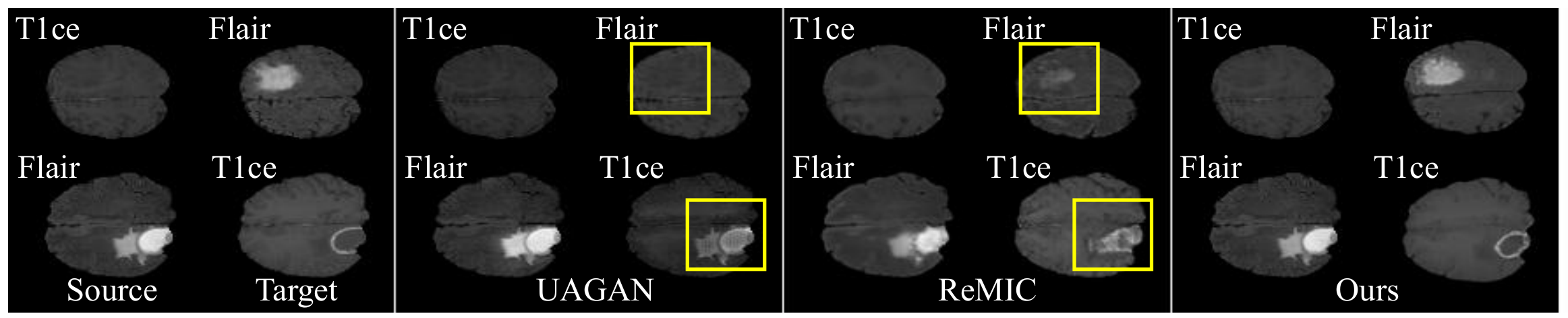} 
	\caption{Image translation results (from source to target) between T1ce and Flair. Yellow boxes highlight the failed translation of brain tumors. In each image, reconstructed images are in the left column, and translated images are in the right column.}
	\label{fig5}
\end{figure*}

\begin{figure*}[t]
	\centering
	\includegraphics[width=0.8\textwidth]{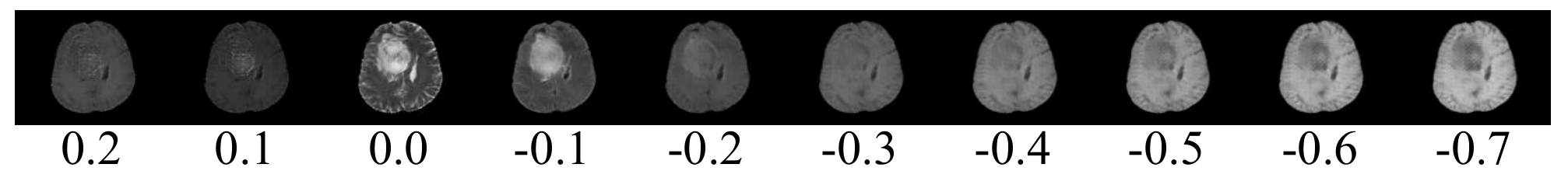} 
	\caption{Evaluation of the effect of the style codes. Images of each column are generated by interpolating the values of 3rd dimension with the rest fixed.}
	\label{fig8}
\end{figure*}

\subsubsection{Image translation}
Since the Flair is the most informative modality for the segmentation of WT, and T1ce is for the segmentation of TC and ET~\cite{hgcb:16,zcvr:21}, we discuss the results of image translation between these two modalities in Fig.~\ref{fig5}. Note that the translation performances are only evaluated against UAGAN and ReMIC, since there is no translation for nnU-Net. For the translation between modalities T1ce and Flair that convey different biological information (Fig.~\ref{fig1}, for example) of brain tumors, our model is superior to others, particularly for the tumor areas translation. Furthermore, as shown in Table~\ref{table4}, our model outperforms other methods on translation task in terms of SSIM, which suggests that our model can produce more realistic images, and more effectively exploit accurate complementary information to improve segmentation.

\subsubsection{Ablation study}
In this section, we assess the contribution of different components in WT segmentation on BraTS2020. We denote the \textbf{C}ontent \textbf{c}onsistency loss (Eq.~\eqref{loss:cc_inter}), the inter-modality \textbf{T}ranslation loss (Eq.~\eqref{loss:t_inter}), \textbf{F}eature disentanglement framework and \textbf{C}urriculum \textbf{d}isentanglement learning as Cc, T, F and Cd, respectively. As shown in Table~\ref{tableablation}, we describe the ablation experiments as follows: (1) \textbf{Ours w/o Cc} denotes that our model is trained without the content consistency loss. (2) \textbf{Ours w/o T} denotes that our model is trained without the inter-modality translation loss. (3) \textbf{Ours w/o F} denotes that the style encoder $ E_{s} $ and image generation decoder $ D_{t} $ are deactivated. In this experment, our model is trained without feature disentanglement. Since Cc, T and Cd are based on feature disentanglement framework, these components cannot be applied in Ours w/o F, and we denote it as \textbf{Ours w/o Cc, T, Cd, F} in Table~\ref{tableablation}. (4) \textbf{Ours w/o Cd} denotes that we train our model only at inter-modality step for all 50 epochs. Table~\ref{tableablation} shows that best results are achieved with all components. The performance is significantly improved from 78.36\% (\textbf{Ours w/o Cc, T, Cd, F}) to 82.91\%. In addition, the model can also be trained with both style-consistent learning and inter-modality learning for 50 epochs in an end-to-end manner (\textbf{Our end-to-end}). Compared with the two-step training scheme, the average Dice score dropped from 82.91\% to 82.30\%. 

\begin{figure}[t]
	\centering
	\includegraphics[width=0.8\columnwidth]{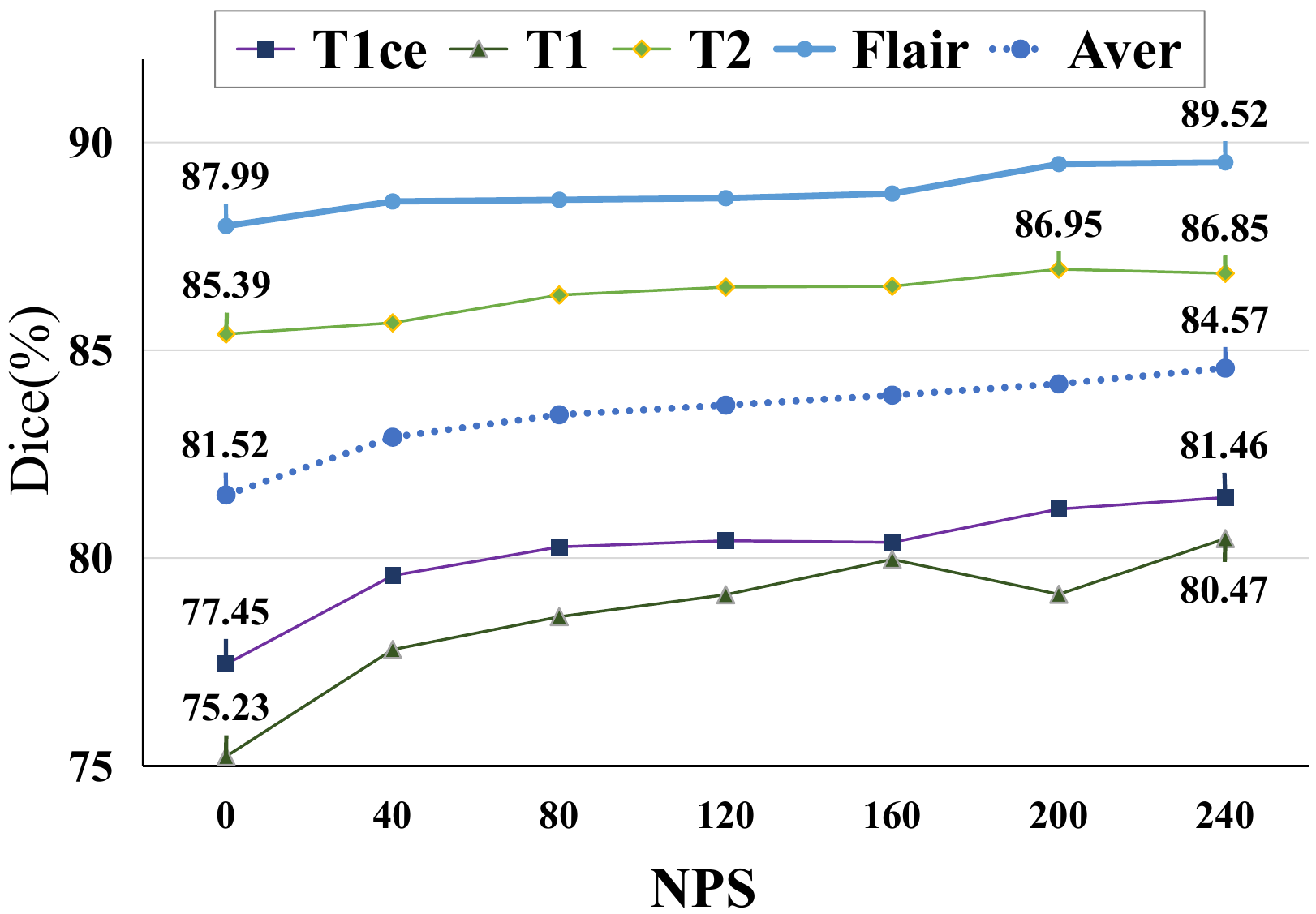}
	\caption{The ratio test for the WT segmentation task on BraTS2020. Only the header and tail values as well as the best values are displayed. NPS: the Number of Paired Subjects.}
	\label{fig7}
\end{figure}

\subsubsection{Influence of paired subjects}
We conduct a ratio test to investigate the effect of the paired subjects. We keep the number of training subjects fixed, and assign different numbers of subjects as paired data. As shown in Fig.~\ref{fig7}, introducing paired data in training does improve the performance of our model. Note that our model is still better than the state-of-the-art methods when the Number of Paired Subjects (NPS) equals 0, and can get satisfactory results when NPS is relatively small (40, for example), which is a good news for clinical practice. 

\subsubsection{Disentanglement evaluation}
We qualitatively examine the effect of each dimension of style code $ s $ with latent space arithmetics~\cite{cpws:20} on 10 subjects. We set the style code size to $ n_{s}=8 $ as suggested by related work ~\cite{cpws:20,lthsy:18}. We conduct statistical analysis on style codes obtained from all the test images, and the max, min and avearge values are 0.189, -0.678 and -0.014, respectively. Note that, we use L2 norm regularization to constrain the style codes. Therefore, interpolating in the range [-0.7, 0.2] covers the possible space. We discover that image style are controlled by the 3rd dimension. As shown in Fig.~\ref{fig8}, images of each column are generated by interpolating the values of 3rd dimension with the rest fixed. In addition, we change the value of the 3rd dimension with others fixed, and compare the synthetic images with the corresponding four real images. The SSIM values for T1ce, T1, T2 and Flair get the maximum of 0.7203, 0.7017, 0.6668 and 0.6963 when the value of 3rd dimension is set to 0.1, -0.3, 0.0 and -0.1, respectively. 

\section{Conclusion}
In this paper, we propose a novel framework that leverages privileged semi-paired images for multi-modal brain tumor segmentation. Specifically, we develop a two-step curriculum disentanglement learning model which can be trained with semi-paired images and make predictions with unpaired images as inputs. The style and content of input images are extracted separately through the two steps. Furthermore, with limited paired images, we incorporate the supervised translation and content consistency loss to enhance the exploitation of the encoded complementary information. Both the quantitative and qualitative evaluations show the superiority of our proposed model in comparison with the state-of-the-art methods. We will further study brain tumor segmentation in a semi-supervised and privileged semi-paired learning setting.

\section*{Acknowledgment}
This work is supported in part by the Guangzhou Science and Technology Planning Project (202201010092), the Guangdong Provincial Natural Science Foundation (2020A1515010717), NSF-1850492 and NSF-2045804.



\end{document}